\documentclass[letterpaper, 10 pt, conference]{ieeeconf}  

\IEEEoverridecommandlockouts                              

\usepackage{times}
\usepackage[table]{xcolor}
\usepackage{soul}
\usepackage[utf8]{inputenc}
\usepackage{caption}

\usepackage{comment}
\usepackage{epsfig}
\usepackage{epstopdf}
\usepackage{graphics}
\usepackage{subfigure}
\usepackage{amsmath}
\usepackage{amssymb} 
\usepackage{xspace}
\usepackage{multirow}
\usepackage{hyperref}
\usepackage{float}
\usepackage{balance}

\usepackage{siunitx,  
            booktabs, 
            caption}  
            
\definecolor{lightgray}{gray}{0.85}

\newif\ifcommenton
\commentontrue
\ifcommenton
\newcommand{\red}[1]{\textcolor{red}{#1}}
\newcommand{\blue}[1]{\textcolor{blue}{#1}}
\else
\newcommand{\red}[1]{}
\newcommand{\blue}[1]{}
\fi

\newcommand{\ourgan}{\mbox{SC-GAN}\xspace}
\newcommand{\noscenebaselinegan}{\mbox{GAN-no-scene}\xspace}
\newcommand{\baselinegan}{\mbox{GAN-concat-scene}\xspace}
\newcommand{\noscenelstmbaselinegan}{\mbox{GAN-LSTM-no-scene}\xspace}
\newcommand{\lstmbaselinegan}{\mbox{GAN-LSTM-concat-scene}\xspace}

\title{Improving Movement Predictions of Traffic Actors in Bird's-Eye View\\ Models using GANs and Differentiable Trajectory Rasterization}

\author{Eason Wang\textsuperscript{*}\thanks{\textsuperscript{*}Authors contributed equally. This work was done while
Eason Wang was an intern at Uber ATG in Pittsburgh, PA, USA.}, Henggang Cui\textsuperscript{*}, Sai Yalamanchi, Mohana Moorthy, Fang-Chieh Chou, Nemanja Djuric \\
  Uber Advanced Technologies Group\\
  \small{\texttt{\{junheng, hcui2, syalamanchi, mmoorthy, fchou, ndjuric\}@uber.com} }\\
}

\begin{document}
\maketitle
\thispagestyle{empty}
\pagestyle{empty}


\begin{abstract}
One of the most critical pieces of the self-driving puzzle is the task of predicting future movement of surrounding traffic actors, which allows the autonomous vehicle to safely and effectively plan its future route in a complex world. 
Recently, a number of algorithms have been proposed to address this important problem, spurred by a growing interest of researchers from both industry and academia. 
Methods based on top-down scene rasterization on one side and Generative Adversarial Networks (GANs) on the other have shown to be particularly successful, obtaining state-of-the-art accuracies on the task of traffic movement prediction. 
In this paper we build upon these two directions and propose a raster-based conditional GAN architecture, powered by a novel differentiable rasterizer module at the input of the conditional discriminator that maps generated trajectories into the raster space in a differentiable manner. 
This simplifies the task for the discriminator as trajectories that are not scene-compliant are easier to discern, and allows the gradients to flow back forcing the generator to output better, more realistic trajectories. 
We evaluated the proposed method on a large-scale, real-world data set, showing that it outperforms state-of-the-art GAN-based baselines.

\end{abstract}

\section{Introduction} 
\label{sect:introduction}

Progress in Artificial Intelligence (AI) applications has experienced strides in the last decade, with a number of highly publicized success stories that have captured attention and imagination of professionals and laymen alike. 
Computers have reached and even surpassed human performance in centuries old games such as go and chess \cite{silver2016mastering,silver2017mastering}, are starting to understand health conditions and suggest medical treatments \cite{topol2015patient}, and can reason about complex relationships conveyed through images \cite{vinyals2017show}. 
Prior to this, AI was already well-established at the very core of multibillion-dollar industries such as advertising \cite{broder2008search} and finances \cite{nuti2011algorithmic}, where today algorithmic approaches have all but replaced human experts. 
Nevertheless, despite these advancements the AI revolution is far from over \cite{harari2017reboot}. 
The automobile domain is one of the last major industries to be disrupted by the current wave, where AI is yet to make its greatest impact through the development and deployment of self-driving vehicles (SDVs). 
Coupled with the latest breakthroughs in hardware and the recent advent of electric vehicles at a larger scale, this opens doors to potentially redefine our cities and our very way of life.

Although seemingly slow to adopt new technologies, today's automotive and transportation companies are embracing the AI and making extensive use of advanced algorithms \cite{gusikhin2007intelligent}. 
The scope of application of the AI ranges from optimizing the production pipelines to help reduce costs and improve manufacturing procedures \cite{hofmann2017artificial}, to connecting vehicles into networks \cite{singh2015internet} and developing wide-scale intelligent transportation systems to improve overall traffic efficiency \cite{qureshi2013survey}.
Most importantly, a particular focus of the automakers has been on safety of passengers and other traffic actors, defined as other active participants in traffic such as vehicles, pedestrians, and bicyclists.
As a result, recent years have seen a surge of advanced driver-assistance systems (ADAS) \cite{shaout2011advanced}, including advanced cruise control, collision avoidance, or lane departure systems, to name a few. 
While advancing safety on our roads, these systems are only mitigating the unreliable human factor that is the main cause of a vast majority of traffic accidents \cite{nhtsa2018reasons}, and there is a lot to be done to improve road fatalities statistics that are still among the worst in the past decade \cite{nhtsa2018motorvehicle}. 
A possible solution to this concerning situation is a development of the self-driving technology, which holds promise to completely remove the human factor from the equation, thus improving both safety and efficiency of the road traffic.
The efforts on this technology started several decades ago \cite{pomerleau1989alvinn,pomerleau1992neural}, however only recently has it become a center of attention for researchers from both industry and academia \cite{urmson2008self,Bertha2015}.

One of the key factors and requirements for the autonomous driving technology to be safely deployed in complex urban environments are efficient and accurate detection, tracking, and motion prediction of surrounding traffic actors \cite{urmson2007tartan,reinholtz2009odin}. 
In the current work we focus on the problem of prediction, tasked with capturing and inferring accurate and realistic future behavior and uncertainty of actor movement, in order to ensure more efficient, effective, and safe SDV route planning.
This is a critical component of the autonomous system, and significant progress has been made along that direction in recent years. 
In particular, the current state-of-the-art approaches are bird's-eye view (BEV) rasterization methods \cite{dp2018,luo2018fast,cui2019multimodal} on one side that generate BEV raster images of an actor's surroundings as an input to deep networks (see Figure \ref{fig:architecture} for an example image), and Generative Adversarial Networks (GANs) \cite{gupta2018social,sadeghian2019sophie,kosaraju2019social} on the other. 
When combined, these two approaches provide all the context required for the task, while matching a real-world distribution through the adversarial framework. 
However, the output trajectories are still far from optimal, oftentimes not fully obeying the physical and map constraints present in the scene. 
We address this important issue, and propose a novel GAN architecture conditioned on an input raster image, referred to as Scene-Compliant GAN (SC-GAN). 
The critical component of the architecture is the differentiable rasterizer, which allows projection of predicted trajectories directly into the raster space in a differentiable manner. 
This simplifies the discriminator task, leading to higher efficiency of the adversarial training and more realistic output trajectories, as exemplified in the evaluation section.

Main contributions of our work are summarized below:
\begin{itemize}
  \item we present a raster-based SC-GAN architecture based on a novel differentiable rasterizer module, simplifying the discriminator's job while allowing the gradients to flow back to the generator;
  \item we evaluate the proposed method on a large-scale, real-world data, showing that SC-GAN outperforms the existing state-of-the-art generative approaches.
\end{itemize}

\section{Related work}
\label{sect:related_work}
In this section we give an overview of the relevant literature on the topic of traffic motion prediction. 
We first discuss approaches focused on providing scene-compliant trajectories, followed by a discussion on recently proposed state-of-the-art methods based on adversarial training.

\subsection{Predicting scene-compliant trajectories}
Trajectory prediction of surrounding actors on the road is a critical component in many autonomous driving systems \cite{CosgunMCHDALTA17,bansal2018chauffeurnet}, as it allows the SDV to safely and efficiently plan its path through a dynamic traffic environment. 
A big part of the solution to this problem is ensuring that the predicted trajectories are compliant with the given scene, obeying the constraints imposed by the other actors as well as the map elements. 
The method described by Mercedes-Benz \cite{Bertha2015} makes direct use of the mapped lane information, associating actors to the nearby lanes and predicting trajectories along the lane geometry. 
The proposed method does not require learning and is based on heuristics, which may be suboptimal for uncommon traffic scenarios (such as in the case of non-compliant actor behavior).

Beyond using map info to predict behavior through hand-tuned rules, recently a number of methods were proposed that use the map data and information on dynamic objects in SDV's vicinity as inputs to learned algorithms.
The current state-of-the-art approaches rasterize static high-definition map and the detected objects in BEV raster images, ingested by deep networks trained to predict future trajectories \cite{luo2018fast,cui2019multimodal,chou2018predicting,chai2019multipath}.
The raster images are used to convey \emph{scene context information} to the model, while convolutional neural networks (CNNs) are commonly employed to extract \emph{scene context features} from the context image. 
Then, a trajectory decoder module is used to generate trajectory predictions based on the computed scene context features and the actor's past observations.
The scene context info used by the model represents a strong prior for behavior prediction. 
This especially holds true for vehicle actors, and allows the network to learn to predict \emph{scene-compliant} trajectories that follow lanes and obey traffic rules \cite{dp2018,cui2019multimodal,hong2019rules}. 

Despite the benefits of rasterized input representation, the learned models may still output non-compliant trajectories as the predictions are not constrained in any way. 
Authors of \cite{yalamanchi2020itsc} proposed to combine learned and hand-tuned approaches to constrain the output trajectories.
There are several published works on directly improving the scene compliance of learned models through extensions of a loss function.
In ChauffeurNet~\cite{bansal2018chauffeurnet}, DRF-NET~\cite{jain2019discrete}, and in the work by Ridel et al.~\cite{ridel2019scene}, the authors proposed to predict an occupancy probability heatmap for each prediction horizon, where they explicitly penalized the probability mass in off-road regions of the scene to enforce scene compliance.
Niedoba et al.~\cite{niedobaimproving} proposed a scene-compliance loss applicable to models that directly predict trajectory point coordinates, as opposed to predicting occupancy probability heatmaps.
In addition, they also proposed novel scene-compliance metrics that quantify how often the trajectories are predicted to go off-road.
In this work, instead of manually designing losses to penalize non-compliant outputs, we leverage the GAN framework and train a model in an adversarial fashion to encourage more realistic, scene-compliant trajectories.

\begin{figure*}[th!]
   \centering
   \includegraphics[keepaspectratio=1,width=2.0\columnwidth]{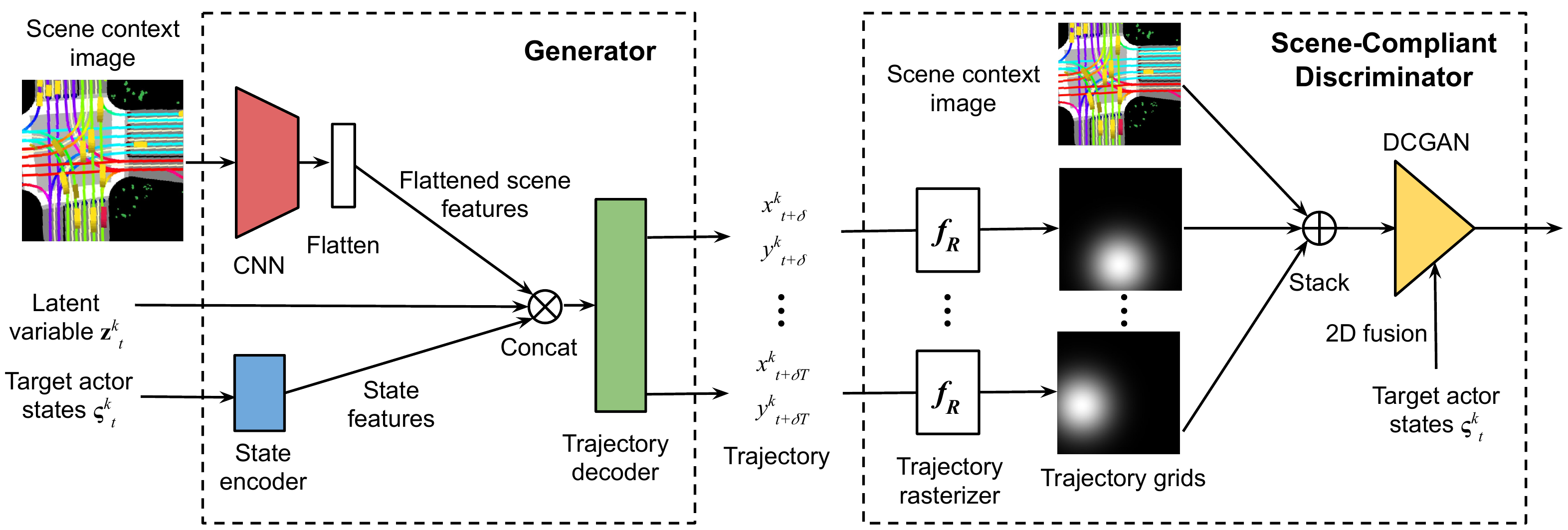}
    \caption{The proposed \ourgan architecture; modules marked in color are learned during training}
    \label{fig:architecture}
\end{figure*}

\subsection{GAN-based trajectory predictions}
Approaches based on GANs have shown outstanding performance in a variety of machine learning tasks \cite{goodfellow2014generative,dai2017towards,zhang2017adversarial}. 
Following the success of general GAN architectures, a number of studies applying adversarial models to trajectory prediction have been proposed \cite{gupta2018social,sadeghian2019sophie,kosaraju2019social,zhao2019multi}.
A common theme in these works is that the generator network outputs trajectory predictions given the inputs, while the discriminator network classifies whether the provided trajectory is coming from the generator or from the ground truth.
The gradients from the discriminator help the generator push the distribution of the generated trajectory predictions closer to the ground-truth distribution.
However, most of these GAN-based models do not condition on the scene context image in the discriminator (e.g., Social-GAN~\cite{gupta2018social}, Sophie~\cite{sadeghian2019sophie}, MATF~\cite{zhao2019multi}), leading to suboptimal performance.
In particular, the discriminator encodes the input trajectory with a Long Short-Term Memory (LSTM) encoder and makes classifications based solely on the trajectory embeddings.
As a result, the discriminator is not able to distinguish between an actual ground-truth trajectory and a trajectory close to the ground truth that is not scene-compliant.

In a recent work \cite{kosaraju2019social} the authors proposed Social-BiGAT that addressed this issue by including the scene context image as an input to the discriminator.
Their discriminator uses a CNN to extract features from the scene context image, and simply concatenates the flattened scene context features with the trajectory embeddings in order to perform classification.
This approach, however, will result in the trajectory embeddings being generated in a separate path from the scene context embeddings, and a non-compliant prediction (e.g., trajectory going off-road) may not trigger a strong activation in the discriminator network.
On the other hand, the proposed \ourgan method has a novel Scene-Compliance Discriminator architecture that transforms the trajectory input into a sequence of 2D occupancy grids in a differentiable way, and stacks the occupancy grids with the scene context image along the channel dimension.
By combining the projected trajectory with the scene context image in the same raster space, the discriminator becomes more sensitive to non-compliant trajectories.
Evaluation on real-world data collected on public roads shows that the novel model design makes the predicted trajectories considerably more scene-compliant and accurate, compared to \mbox{Social-BiGAT} and other state-of-the-art generative approaches.

\section{Scene-Compliant GAN (\ourgan)}
\label{sect:methodology}

In this section we propose a GAN-based method aimed at producing trajectories that better follow constraints existing within the scene. 
We first describe the problem setting, followed by the introduction of generator and discriminator submodules, as well as the novel differentiable rasterizer. 
Lastly, we discuss the loss used during training.

\subsection{Problem statement}

Let us assume that we have access to a real-time data stream coming from sensors such as lidar, radar, or camera, installed onboard a self-driving vehicle. 
In addition, we assume that this data is used as an input by an existing detection and tracking system, outputting state estimates for all surrounding actors up to the current time $t_c$. 
The tracked state comprises the bounding box, center position $[x, y]$, velocity $v$, acceleration $a$, heading $\theta$, and heading change rate $\dot{\theta}$, where the tracker provides its outputs at a fixed sampling frequency of $10Hz$, resulting in the discrete tracking time step of $\delta = 0.1s$.
We denote state output of the tracker for the $i$-th actor at time $t$ as ${\bf s}^i_t = [x^i_t, y^i_t, v^i_t, a^i_t, \theta^i_t, \dot{\theta}^i_t]$, and the total number of actors tracked at time $t$ as $N_t$ (note that in general the actor counts vary for different time steps as new actors appear within and existing ones disappear from the sensor range).
Moreover, we further assume access to a detailed, high-definition map information of the SDV's operating area denoted by $\mathcal{M}$, including road and crosswalk locations, lane directions, and other relevant map information such as observed traffic lights and signage.

Let $\mathcal{S}_{t_c} = \{{\bf S}_{t_c - (L-1) \delta}, {\bf S}_{t_c - (L-2) \delta}, \ldots, {\bf S}_{t_c}\}$ denote the tracked states of all actors detected at time $t_c$ over the past $L$ timestamps, where ${\bf S}_{t} = \{{\bf s}^1_t, {\bf s}^2_t, \dots, {\bf s}^{N_t}_t\}$ represents the state of all actors at time $t$. Then, given the state information $\mathcal{S}_{t_c}$ and the map data $\mathcal{M}$, our goal is to predict future positions of a target actor $k$ over the next $T$ timestamps ${\bf o}^k_{t_c} = [x^k_{t_c+\delta}, y^k_{t_c+\delta}, \dots, x^k_{t_c+T\delta}, y^k_{t_c+T\delta}]$, with $k$ being in the $[1, N_{t_c}]$ range.
The output trajectory is in the actor frame of reference, with origin at the center position, $x$-axis defined by the actor's heading, and $y$-axis defined by the left-hand side of the actor.
Without loss of generality, and similarly to the existing trajectory prediction work \cite{dp2018,cui2019multimodal,bansal2018chauffeurnet,chou2018predicting,chai2019multipath,hong2019rules}, we only predict future center positions of the actor, from which other state variables can be determined. 
Alternatively, one could extend the proposed model to predict actor's full future states ${\bf s}^k_{t_c}$ by using ideas proposed in~\cite{cui2019deep}, however this is beyond the scope of our current work.

\subsubsection{Scene rasterization}
For each individual actor we separately predict future trajectories, where the input to the network is generated through rasterization approaches similar to those used in earlier work \cite{dp2018,cui2019multimodal,bansal2018chauffeurnet,chou2018predicting,chai2019multipath,hong2019rules}. 
In particular, for the $k$-th actor the scene context information on the surrounding tracked actors and the map constraints are provided to the network by rasterizing the map $\mathcal{M}$  and all actors' past polygons from $\mathcal{S}_{t_c}$ onto a per-actor RGB raster image $\mathcal{I}^k_{t_c}$ with resolution $r$. 
The raster can be represented by a matrix of size $H \times W \times 3$, with the actor of interest $k$ located at cell index $[h_0, w_0]$ and rotated such that actor's heading is pointing up.
In our experiments we set $H=W=300$, $h_0 = 50$, $w_0 = 150$, with $r=0.2$ meters per pixel, such that the raster image captures $50m$ in front of the actor, $10m$ behind, and $30m$ on both sides of the actor.
The rasterized map elements include road polygons, driving paths and their directions, crosswalks, detected traffic signals, and other relevant info. 
Actors' polygons are rasterized with different shadings representing different past timestamps, and the target actor is colored differently from other actors in its surroundings (see Figure \ref{fig:architecture} for an example, where we used red and yellow colors to differentiate them, respectively).

\begin{figure*}[t!]
    \centering
    \includegraphics[keepaspectratio=1,width=\columnwidth]{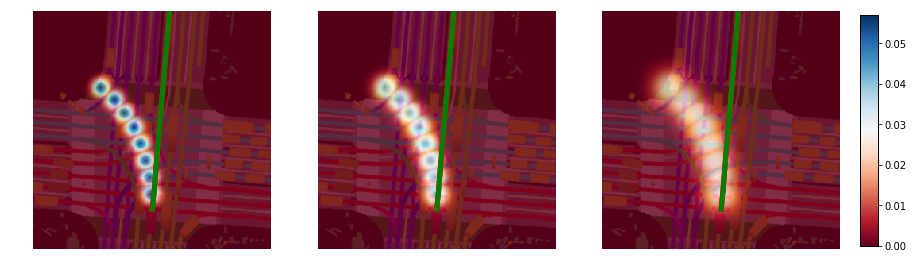} 
    \includegraphics[keepaspectratio=1,width=\columnwidth]{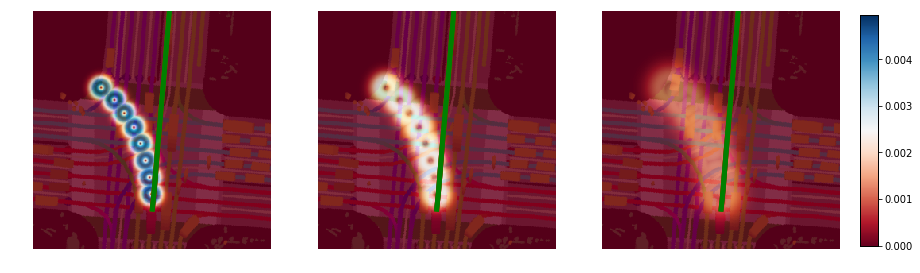}
    \caption{
        Visualization of the differentiable rasterizer by varying $\sigma \in \{1.4, 2, 3\}$~meters, respectively, with non-scene-compliant predicted trajectory (going into opposite lane and off-map) overlaid onto the scene and the ground-truth trajectory shown in green; {\bf left}: rasterizer output;  {\bf right}: gradient norms;
        \normalsize{
            with larger $\sigma$ the gradients are more spread out but are also weaker, while
            for smaller $\sigma$ values the gradients are stronger but are more concentrated
            (e.g., a network using such configuration might not learn effectively if the rasterized point is too far away from the ground truth or the on-road map elements)
        }
    }
    \label{fig:rasterviz}
\end{figure*}

\subsection{Model architecture}

In this section we present the overall architecture of the proposed model, illustrated in Figure~\ref{fig:architecture}. 
It consists of three main modules: 1) generator network, 2) discriminator network, and 3) differentiable trajectory rasterizer.
In the following we describe each module in more detail.

\subsubsection{Generator network}
\label{sect:gen_network}
The generator network $G$ (parameterized by the parameter set ${\boldsymbol \theta}_G$) generates the trajectory prediction $\hat{\bf o}^k_{t_c}$ given the concatenated actor's state inputs ${\boldsymbol \varsigma}^k_{t_c} = [{\bf s}^k_{t_c - (L-1)\delta}, \ldots, {\bf s}^k_{t_c}]$, per-actor raster $\mathcal{I}^k_{t_c}$, and a noise vector ${\bf z}^k_{t_c}$ of dimensionality $d$ with each element sampled from a normal distribution $\mathcal{N}(0,1)$,
\begin{equation}
\hat{{\bf o}}_{t_c} = G({\boldsymbol \varsigma}_{t_c}, \mathcal{I}_{t_c}, {\bf z}_{t_c}; {\boldsymbol \theta}_G),
\end{equation}
where here and in the remainder of the section we drop the actor index superscript $k$ to simplify the notation.
As can be seen in Figure \ref{fig:architecture}, the generator first extracts the scene context features from the scene image $\mathcal{I}_{t_c}$ using a convolutional neural network. 
While any CNN can be used for this purpose \cite{dp2018}, in order to allow for fast real-time inference onboard the SDVs we used MobileNet~\cite{sandler2018inverted}.
Past observed actor states ${\boldsymbol \varsigma}_{t_c}$ are also embedded with a shallow encoding layer, and concatenated with the extracted scene context features and the latent noise vector before being passed to a trajectory decoder module that generates the trajectory predictions.

\subsubsection{Discriminator network}

The discriminator network (parameterized by the parameter set ${\boldsymbol \theta}_D$) 
classifies whether a given future trajectory ${\bf o}_{t_c}$ is coming from a ground truth (i.e., {\it true}) or the generator (i.e., {\it fake}), conditioned on the past observed states ${\boldsymbol \varsigma}_{t_c}$ and the scene context image $\mathcal{I}_{t_c}$.
In the previous GAN-based work, either the discriminator architectures did not use the scene context information at all
(e.g., Social-GAN~\cite{gupta2018social}, Sophie~\cite{sadeghian2019sophie}, and MATF~\cite{zhao2019multi}),
or the features from the scene context image $\mathcal{I}_{t_c}$ and from the provided trajectory ${\bf o}_{t_c}$ were extracted using two separate networks, followed by a simple concatenation of the computed features (e.g., Social-BiGAT~\cite{kosaraju2019social}).
Such discriminator design is not sufficiently sensitive to non-compliant trajectories due to the separation of handling of the two inputs, as confirmed in our experiments.
In addition, these discriminators require the use of fully-connected layers, which is advised against by some authors~\cite{radford2015unsupervised}.
In this work we propose a scene-compliant architecture that is more sensitive to non-compliant trajectories, comprising only fully convolutional layers. 
The proposed discriminator relies on a novel module called differentiable trajectory rasterizer, described in the following section.

\subsubsection{Differentiable rasterizer}

The trajectory rasterization module of the scene-compliant discriminator is tasked with rasterizing the future trajectory ${\bf o}_{t_c}$ (either predicted or ground-truth) into a sequence of 2D occupancy grids $\Gamma_{t_c} = \{\mathcal{G}_{t_c+\delta}, \ldots, \mathcal{G}_{t_c+T\delta}\}$,
where each $\mathcal{G}_t = f_R([x_t, y_t], \sigma)$ encodes a single trajectory point at time $t$ and is an $H \times W$ 2D grid with the same shape and resolution as the raster image $\mathcal{I}_{t_c}$,
$f_R(\cdot)$ is a rasterizer function, $[x_t, y_t]$ are coordinates of the trajectory point to be rasterized, and $\sigma$ is a predefined visualization hyper-parameter to be discussed promptly.

Let us denote distance in the actor frame between a cell $[i, j]$, where $i \in \{1, \ldots, H\}$ and $j \in \{1, \ldots, W\}$, and the trajectory point $[x_t, y_t]$ as follows,
\begin{equation}
\label{eq:delta_raster}
    \boldsymbol{\Delta}_t^{ij} = \big[(i - h_0) r, (j - w_0) r \big] - [x_t, y_t],
\end{equation}
computed by considering the raster resolution $r$ and the origin cell of the actor of interest $[h_0, w_0]$.
Then, the trajectory rasterizer calculates value for cell $[i, j]$ of the image $\mathcal{G}_t$ as a probability density of a 2D Gaussian distribution at $\boldsymbol{\Delta}^{ij}_t$,
\begin{equation}
\label{eq:traj_raster}
    \{\mathcal{G}_t\}_{ij} = \mathcal{N}(\boldsymbol{\Delta}^{ij}_t | {\boldsymbol 0}, {\boldsymbol \Sigma}).
\end{equation}
The covariance matrix is defined as ${\boldsymbol \Sigma} = diag([\sigma^2, \sigma^2])$,
such that the standard deviation $\sigma$ modulates the probability density of the rasterized trajectory point. 
An example of the resulting image is shown in Figure \ref{fig:architecture}.
We let the rasterizer $f_R(\cdot)$ rasterize each trajectory point as a Gaussian occupancy density map, as opposed to a one-hot matrix, as this facilitates back-propagation during training through well-defined gradients. 
In particular, the gradient vector is aligned with the direction of $\boldsymbol{\Delta}^{ij}_t$ and computed as follows,
\begin{equation}
    \label{eq:grad}
\nabla_{[x_t, y_t]} \big(\{\mathcal{G}_t\}_{ij}\big) = [\frac{\partial \{\mathcal{G}_t\}_{ij}}{\partial x}, \frac{\partial \{\mathcal{G}_t\}_{ij}}{\partial y}] = -\frac{\{\mathcal{G}_t\}_{ij}}{\sigma^2} \boldsymbol{\Delta}_t^{ij},
\end{equation}
justifying the name {\it differentiable rasterizer} of the module.

For a fixed point $[x_t, y_t]$ the gradient achieves its maximum $\ell_2$ norm of $1 / (\sqrt{2 \pi e}\sigma^2)$ when $\|\boldsymbol{\Delta}_t^{ij}\|_2=\sigma$ holds.
Let us represent a super-level set of the gradient norm as $\mathcal{S}_\alpha = \{[i, j]: \|\nabla_{[x_t, y_t]} (\{\mathcal{G}_t\}_{ij})\|_2 \geq \alpha\}$, where $0 < \alpha \leq \frac{1}{\sqrt{2 \pi e}\sigma^2}$. 
This set consists of points that form an annulus in the spatial extent of the occupancy grid, and it always contains the ring of points that satisfy $\|\boldsymbol{\Delta}^{ij}_t\|_2=\sigma$.
Points in $\mathcal{S}_\alpha$ will map to gradients of significant magnitudes, as per definition of $\mathcal{S}_\alpha$.
Increasing $\sigma$ will increase the width of the annulus (for a fixed choice of $\alpha$), but will decrease the value of the maximum of a gradient norm.
The choice of $\sigma$ thus controls the balance between how well spread out gradients spatially are and their norm; see Figure~\ref{fig:rasterviz} for visualization of this phenomenon.
It is interesting to note that when $x_t$ or $y_t$ are outside of the 2D grid range $H \times W$ (e.g., this can happen for a particularly poor initialization of the model weights), only a small part of $\mathcal{S}_\alpha$ needs to be contained within the grid in order for $[x_t, y_t]$ to still receive a meaningful gradient.

Once the trajectory is rasterized in a differentiable manner, all 2D occupancy grids $\Gamma_{t_c}$ for a trajectory ${\bf o}_{t_c}$ are stacked together with the scene context image $\mathcal{I}_{t_c}$ along the channel dimension.
This results in a multi-channel image with both the scene context and the trajectory ``plotted'' on top of it.
Using the proposed approach the discriminator has an easier task to decide whether or not the generated trajectory is valid and scene-compliant, as the rasterized scene elements and the trajectory are aligned in the raster space. 
This is unlike prior work where the raster and trajectories were simply concatenated together \cite{kosaraju2019social}, which results in a much more difficult task for the discriminator as we confirm in Section \ref{sect:experiments}.
We employ the fully-convolutional DCGAN architecture~\cite{radford2015unsupervised} as our discriminator,
and also fuse the past observed states ${\boldsymbol \varsigma}_{t_c}$ of the actor of interest into the multi-channel raster using the 2D fusion method proposed in~\cite{chou2018predicting}.


\subsection{Training loss}

Unlike previous GAN-based prediction works~\cite{gupta2018social, sadeghian2019sophie, kosaraju2019social}
which used the vanilla cross-entropy loss as their GAN loss,
we used the Wasserstein GAN loss with gradient penalty~\cite{arjovsky2017wasserstein, DBLP:journals/corr/GulrajaniAADC17}, shown to outperform other approaches.
For brevity, let us denote the discriminator network $D({\boldsymbol \varsigma}_{t_c}, \mathcal{I}_{t_c}, \Gamma_{t_c}; {\boldsymbol \theta}_D)$ by $D(\Gamma_{t_c})$. Let $\mathbb{P}_g$ be the generated data distribution, $\mathbb{P}_r$ the true data distribution, and $\mathbb{P}_{\tilde{{\bf o}}_{t_c}}$ the distribution implicitly defined by sampling uniformly along straight lines between pairs of points sampled from $\mathbb{P}_g$ and $\mathbb{P}_r$ distributions. 
The distribution $\mathbb{P}_{\tilde{{\bf o}}_{t_c}}$ is used for computing the gradient penalty term (see \cite{DBLP:journals/corr/GulrajaniAADC17} for more details). 
Then, the GAN-based loss used to train \ourgan is given as
\begin{equation}
\label{eq:gan_loss}
    \begin{split}
        \mathcal{L} = & \mathop{\mathbb{E}}_{\hat{{\bf o}}_{t_c} \sim \mathbb{P}_g} \big(D(\hat{\Gamma}_{t_c})\big) \;- \mathop{\mathbb{E}}_{{\bf o}_{t_c} \sim \mathbb{P}_r} \big(D(\Gamma_{t_c})\big) \; + \\
      & \; \lambda \mathop{\mathbb{E}}_{\tilde{{\bf o}}_{t_c} \sim \mathbb{P}_{\tilde{{\bf o}}_{t_c}}} \Big(\big(\|\nabla_{\tilde{{\bf o}}_{t_c}} D(\tilde{\Gamma}_{t_c})\|_2-1\big)^2\Big),
    \end{split}
\end{equation}
where $\lambda$ is the gradient penalty weight. 
Moreover, one could also include an additional variety $\ell_2$ loss between the generated and the ground-truth trajectory points, defined as a minimum $\ell_2$ error of $K$ randomly generated samples, as commonly done in GAN-based prediction \cite{gupta2018social, sadeghian2019sophie, kosaraju2019social}.

Lastly, the discriminator network is discarded following the training phase, and only the generator is used for model evaluation. 
We used the efficient MobileNet as a generator, as discussed in Section \ref{sect:gen_network}, which allows for fast model inference that is well-suited for running onboard the SDVs.


\begin{table*} [t!]
\centering
\caption{Comparison of \ourgan and the baselines, with the models trained using only the Wasserstein GAN loss}
\label{tab:pred-errors}
{\normalsize
{
  \begin{tabular}{lcccccccccc}
     & \multicolumn{6}{c}{\bf mean over 3} & \multicolumn{2}{c}{\bf min over 3} & \multicolumn{2}{c}{\bf min over 20} \\
    \cmidrule(lr){2-7} \cmidrule(lr){8-9} \cmidrule(lr){10-11}
     & \multicolumn{2}{c}{\bf ${\boldsymbol \ell_2}$ [m]} & \multicolumn{2}{c}{\bf ORD [m]} & \multicolumn{2}{c}{\bf ORFP [\%]} & \multicolumn{2}{c}{\bf ${\boldsymbol \ell_2}$ [m]} & \multicolumn{2}{c}{\bf ${\boldsymbol \ell_2}$ [m]} \\
    {\bf Method} & {\bf Avg} & {\bf @4s} & {\bf Avg} & {\bf @4s} & {\bf Avg} & {\bf @4s} & {\bf Avg} & {\bf @4s} & {\bf Avg} & {\bf @4s} \\
    \hline
    \rowcolor{lightgray}
    \noscenebaselinegan & 4.13 & 6.57 & 0.840 & 1.203 & 24.50 & 30.28 & 3.74 & 5.87 & 3.30 & 5.13 \\
    \baselinegan & {\bf 2.35} & {\bf 5.62} & 0.152 & 0.435 & 4.40 & 12.22 & 1.37 & 3.13 & 0.63 & 1.30 \\
    \rowcolor{lightgray}
    \noscenelstmbaselinegan & 3.44 & 8.30 & 0.793 & 2.703 & 22.54 & 61.02 & 3.22 & 8.04 & 2.97 & 7.76 \\
    \lstmbaselinegan & 3.08 & 6.10 & 0.258 & 0.732 & 6.88 & 17.40 & 1.94 & 4.13 & 1.11 & 2.58 \\
    \rowcolor{lightgray}
    \ourgan & 2.44 & 5.86 & {\bf 0.085} & {\bf 0.204} & {\bf 2.11} & {\bf 5.66} & {\bf 1.29} & {\bf 2.95} & {\bf 0.58} & {\bf 1.20} \\
    \hline
\end{tabular}
}
\vspace{-0.1cm}
}
\end{table*}

\section{Experiments}
\label{sect:experiments}

In this section we present results of empirical evaluation. We first focus on the quantitative comparison of the proposed approach with the current state-of-the-art, followed by an analysis of case studies and the ablation study of the method.

{\bf Baselines:}
We evaluated the following baseline models, focusing on prediction approaches that include the current state-of-the-art social- and/or GAN-based components:

\begin{itemize}
    \item Social-LSTM (S-LSTM), motion prediction method based on LSTM considering social interactions \cite{alahi2016social};
    \item Social-GAN (S-GAN), extension of Social-LSTM that also incorporates the GAN idea~\cite{gupta2018social};
    \item Social-Ways (S-Ways), GAN-based approach that employs Info-GAN and a social attention layer \cite{amirian2019social};
    \item GAN-LSTM-no-scene, where the discriminator uses only trajectory inputs without scene context inputs, and trajectory points are encoded using an LSTM encoder, similar to Sophie~\cite{sadeghian2019sophie} and MATF~\cite{zhao2019multi};
    \item GAN-LSTM-concat-scene, where the discriminator concatenates the scene context features with the trajectory features encoded using an LSTM encoder, similar to Social-BiGAT~\cite{kosaraju2019social};
    \item \noscenebaselinegan and \baselinegan, variants of the above two baselines where the trajectory points are encoded using a fully-connected layer in the discriminator instead of the recurrent architecture;
    \item \ourgan: the proposed model trained with just the GAN loss as given in Equation~\eqref{eq:gan_loss};
    \item \ourgan-$\ell_2$: the proposed model trained with both GAN loss and a variety $\ell_2$ loss with a weight of $10$.
\end{itemize}
We used the available open-sourced code for the baseline approaches Social-GAN\footnote{ \url{github.com/agrimgupta92/sgan}, last accessed Jan. 2020} \cite{gupta2018social}, Social-LSTM\footnote{ \url{github.com/quancore/social-lstm}, last accessed Jan. 2020} \cite{alahi2016social}, and Social-Ways\footnote{ \url{github.com/amiryanj/socialways},  last accessed Jan. 2020} \cite{amirian2019social}.
The proposed \ourgan model and the remaining baselines were implemented in TensorFlow~\cite{tensorflow2015-whitepaper}, using the same generator network but varying the discriminator architectures.
Note that for Social-BiGAT and the other above-mentioned baselines that do not have open-sourced code we used our own implementations of similar architectures.
Unless otherwise mentioned, the implemented baseline models were trained end-to-end from scratch with just the Wasserstein GAN loss, defined in Equation \eqref{eq:gan_loss}.
In order to satisfy the Lipschitz constraint of the Wasserstein GAN loss, we used a gradient penalty weight of $10$.
For the \ourgan-$\ell_2$ model, we computed the variety $\ell_2$ loss by drawing multiple samples (we used three in our experiments) from the generator and using the sample with the lowest $\ell_2$ loss to update the model parameters in the back-propagation phase.
The raster size was set to $300 \times 300$~pixels with the resolution of $0.2m$ per pixel.
For the proposed differentiable rasterizer we set the $\sigma$ parameter from Equation \eqref{eq:traj_raster} to $2m$.
The models were trained with a per-GPU batch size of $64$ and Adam optimizer \cite{kingma2014adam}, the learning rates (for both the generator and the discriminator) were tuned separately for each model using grid search, and we ran three discriminator steps for every generator step to balance the two modules.
Note that for the S-Ways baseline we tried a large number of parameter settings and tweaking the original source code, however the results remained very suboptimal and thus we do not report them in the following sections.

{\bf Data set}
We used a large-scale, real-world ATG4D data set discussed in \cite{meyer2019lasernet}. The data set comprises 240 hours of data obtained by driving in various traffic conditions (e.g., varying times of day, days of the week, in several US cities).
Each actor at each discrete tracking time step (every $0.1s$) amounts to a single data point, which consists of the current and past $0.4s$ of observed actor states (bounding box positions, velocities, accelerations, headings, and turning rates), used as a model input along with the surrounding high-definition map information, and the states for the future $4s$ are used as the ground-truth labels. 
The models were trained to predict trajectory points sampled at $2Hz$ to speed up the training of competing approaches.
After removing static actors, our data set consisted of $7.8$ million data points in total, and we used a $3$/$1$/$1$ split to obtain train/validation/test data sets.

{\bf Evaluation metrics}
The models were evaluated using the standard average $\ell_2$ displacement error (ADE) and final $\ell_2$ displacement error (FDE) metrics, computed over the $4s$ prediction horizon.
For the generative models we drew multiple $K$ samples at inference time and evaluated both the mean and minimum errors over the samples.
The minimum error metric measures the coverage and diversity of the predicted samples and
is a standard approach used in other multimodal prediction work~\cite{cui2019multimodal,gupta2018social, sadeghian2019sophie, kosaraju2019social, chai2019multipath, zhao2019multi}.
In addition to the standard $\ell_2$ metrics, we also evaluated the models using two scene-compliance metrics
introduced in~\cite{niedobaimproving}.
More specifically, for each actor we first identify their likely \emph{drivable regions} by traversing the directed lane graph in the map data starting from the actor's current position,
and define a trajectory point as being \emph{off-road} if it is outside of the drivable region computed in such a way.
Then, the \emph{off-road distance} (ORD) metric measures the distance from a predicted point to the nearest drivable region (defined to be 0 if inside the region), while
the \emph{off-road false-positive} (ORFP) metric measures the percentage of predicted trajectory points that were off-road for cases where the corresponding ground-truth trajectory point was inside the drivable region.

\begin{figure*}[t!]
    \centering
    \includegraphics[keepaspectratio=1,width=0.5\columnwidth]{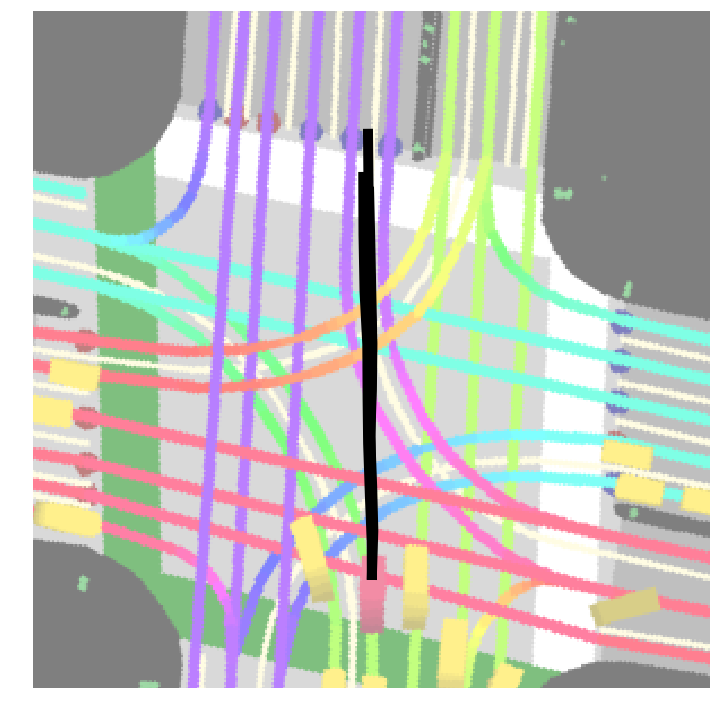}
    \includegraphics[keepaspectratio=1,width=0.5\columnwidth]{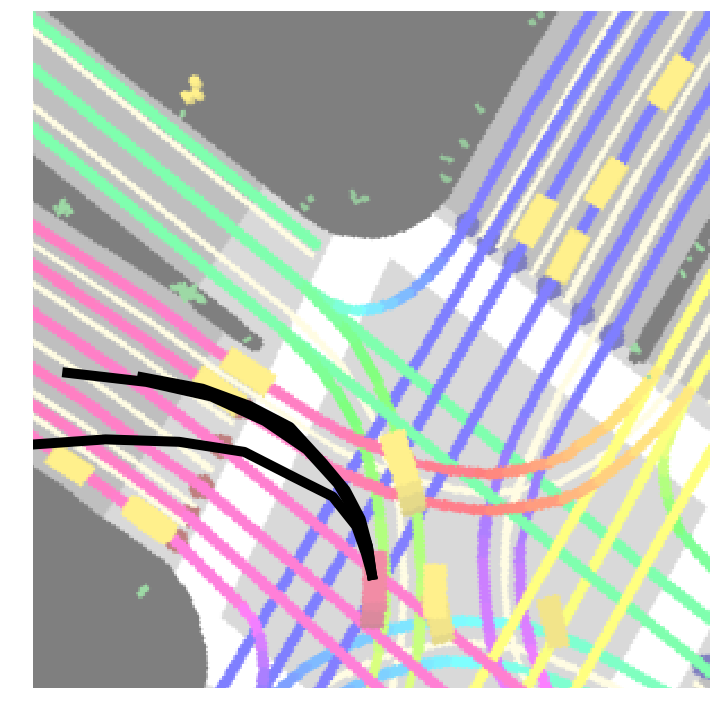}
    \includegraphics[keepaspectratio=1,width=0.5\columnwidth]{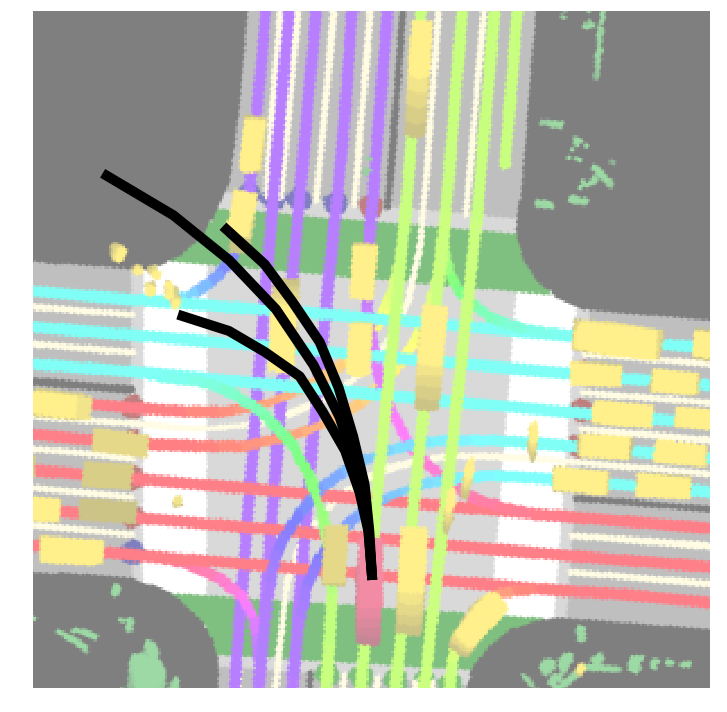}
    \includegraphics[keepaspectratio=1,width=0.5\columnwidth]{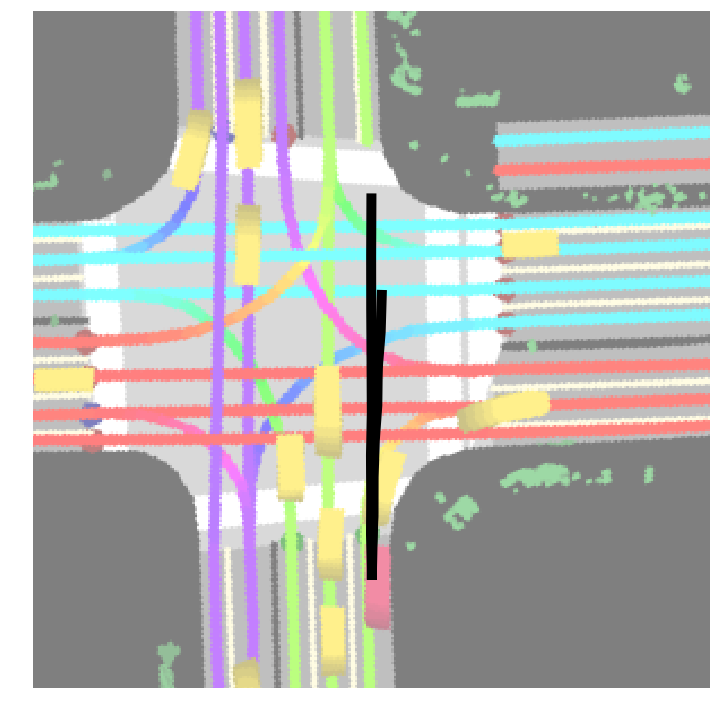} \\
    \includegraphics[keepaspectratio=1,width=0.5\columnwidth]{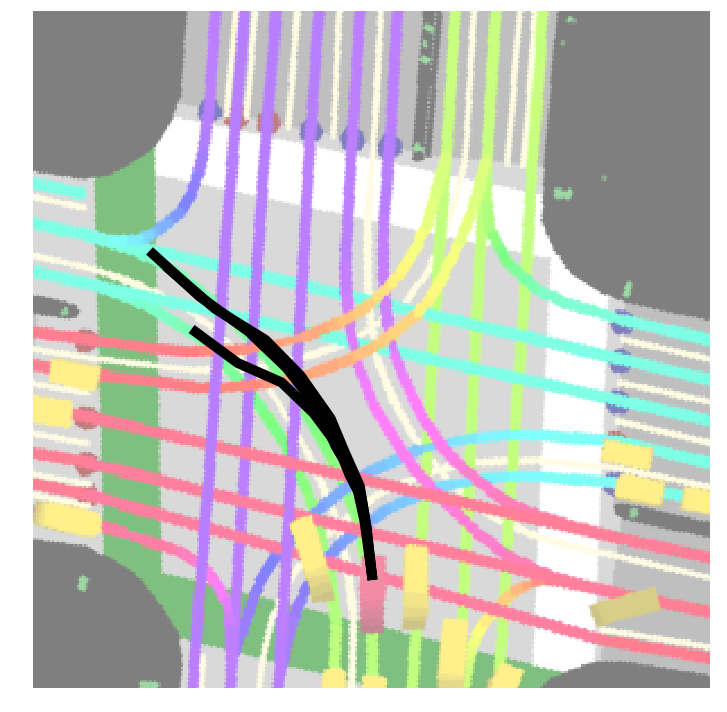}
    \includegraphics[keepaspectratio=1,width=0.5\columnwidth]{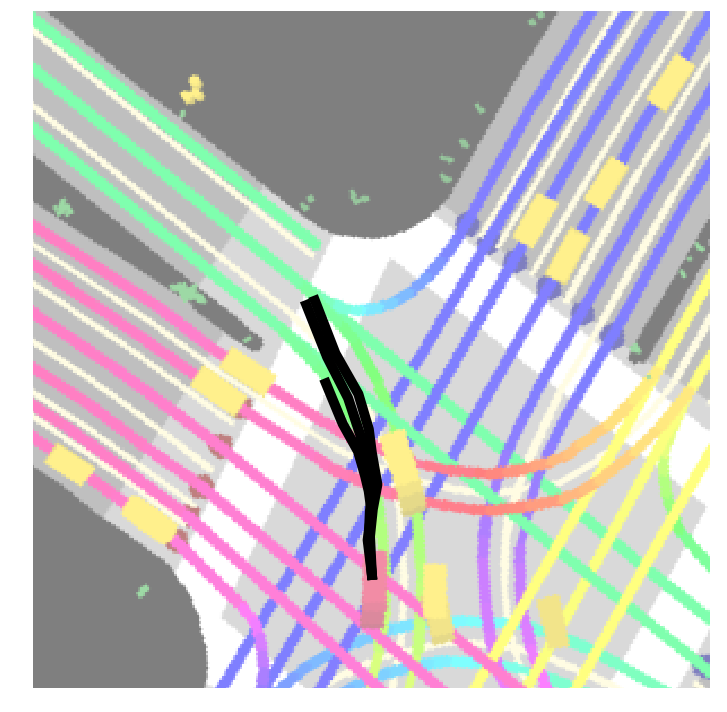}
    \includegraphics[keepaspectratio=1,width=0.5\columnwidth]{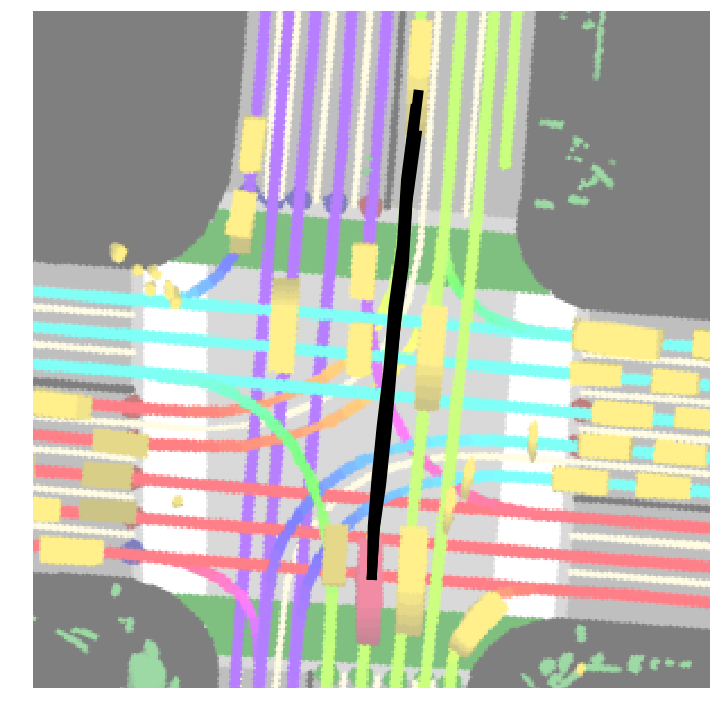}
    \includegraphics[keepaspectratio=1,width=0.5\columnwidth]{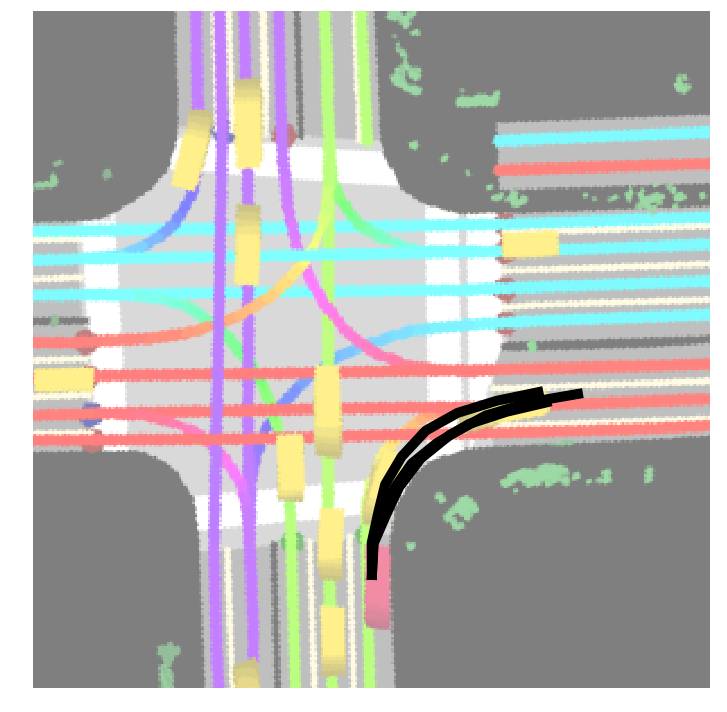}\\
    \includegraphics[keepaspectratio=1,width=0.5\columnwidth]{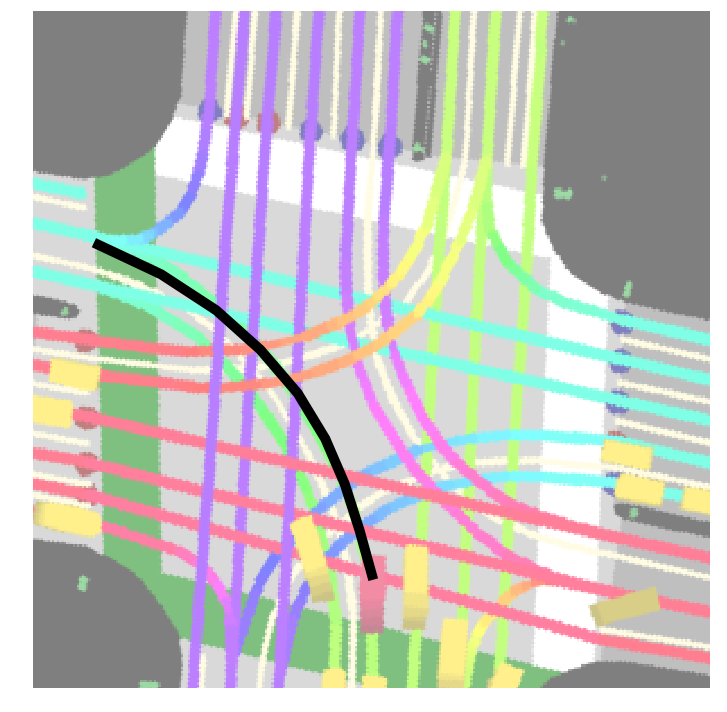}
    \includegraphics[keepaspectratio=1,width=0.5\columnwidth]{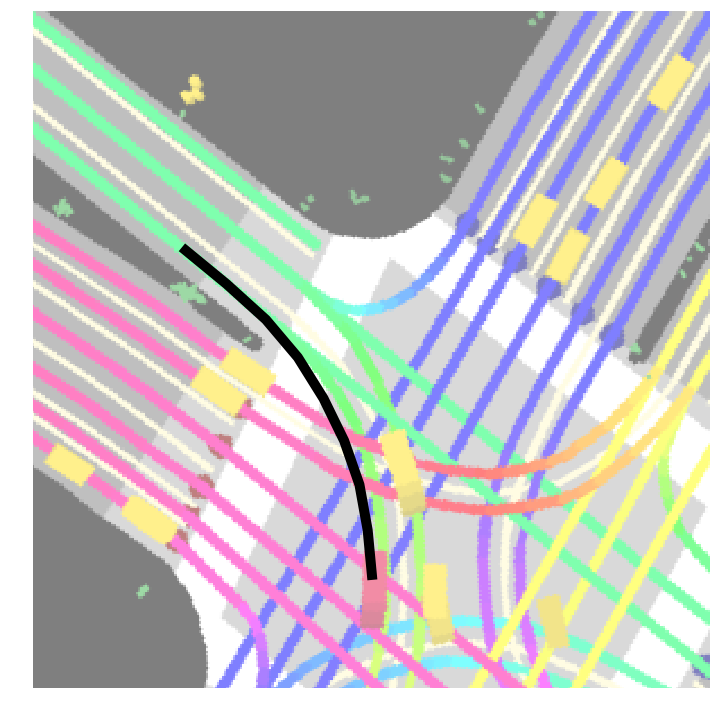}
    \includegraphics[keepaspectratio=1,width=0.5\columnwidth]{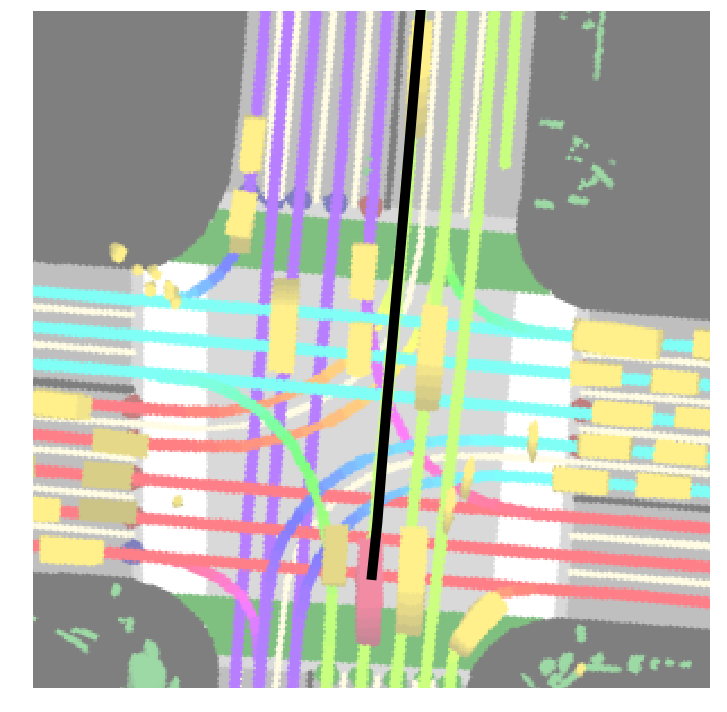}
    \includegraphics[keepaspectratio=1,width=0.5\columnwidth]{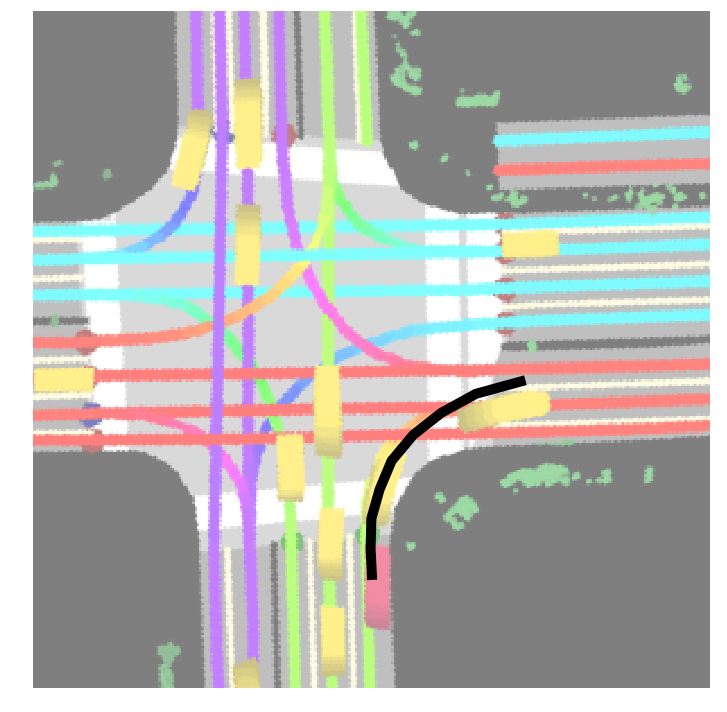} \\
    \caption{
        Qualitative results of GAN architectures conditioned on the raster image with and without using the differentiable rasterizer: {\bf top}: \baselinegan, {\bf middle}: \ourgan, {\bf bottom}: ground-truth trajectories; we show a different traffic scenario in each column, plotting $3$ sampled trajectories for each generative model
    }
    \label{fig:case_studies}
\end{figure*}

\begin{table} [t!]
\centering
\caption{Comparison of $\ell_2$ ADE and FTE error metrics (in meters) of the proposed \ourgan and the state-of-the-art trajectory prediction methods, with the models trained using both GAN and $\ell_2$ losses; note that S-LSTM predicts only a single trajectory so its min-over-$K$ is not reported}
\label{tab:pred-errors_external}
{\normalsize
{
  \begin{tabular}{lcccccc}
     & \multicolumn{2}{c}{\bf mean@3} & \multicolumn{2}{c}{\bf min@3} & \multicolumn{2}{c}{\bf min@20} \\
    \cmidrule(lr){2-3} \cmidrule(lr){4-5} \cmidrule(lr){6-7}
    {\bf Method} & {\bf Avg} & {\bf @4s} & {\bf Avg} & {\bf @4s} & {\bf Avg} & {\bf @4s} \\
    \hline
    \rowcolor{lightgray}
    S-GAN~\cite{gupta2018social} & 3.01 & 7.66 & 2.36 & 5.94 & 1.93 & 4.77 \\
    S-LSTM~\cite{alahi2016social} & 2.93 & 5.17 & - & - & - & - \\
    \rowcolor{lightgray}
    \ourgan-$\ell_2$ & {\bf 1.75} & {\bf 4.17} & {\bf 1.03} & {\bf 2.26} & {\bf 0.54} & {\bf 1.01} \\
    \hline
\end{tabular}
}
\vspace{-0.2cm}
}
\end{table}

\subsection{Quantitative results}

We first compared the performance of methods that use only the GAN loss, in order to evaluate impact of discriminator designs on the GAN approaches in isolation from other effects.
The results of quantitative evaluation of \ourgan and the baselines are presented in Table~\ref{tab:pred-errors}.
Unsurprisingly, \noscenebaselinegan shows very poor results on the two scene-compliance metrics since the discriminator does not make any use of the scene context.
Compared to \noscenebaselinegan, the \baselinegan model improves both the $\ell_2$ errors and the scene-compliance metrics, showing that including the scene context plays a critical role in the discriminator.
We can also see that using an LSTM-based decoder does not lead to improved results, and in fact the opposite conclusion can be made.
Lastly, the proposed \ourgan model had slightly worse mean $\ell_2$ errors compared to \baselinegan,
however its min-over-$K$ $\ell_2$ errors were improved and it also reached significantly better results when considering scene-compliance metrics.
In particular, \ourgan reduced the off-road distance and off-road false positives by over 50\% compared to \baselinegan.
This can be explained by the fact that \ourgan explicitly projects the predicted trajectories into the raster space and stacks them with the scene context image, allowing the discriminator to more effectively identify non-compliant outputs.
Figure~\ref{fig:case_studies} shows several case studies from the validation data illustrating the difference in scene compliance between \ourgan and \baselinegan predictions, discussed in detail in the next section.

In the above analysis we showed that the proposed approach outperformed the existing state-of-the-art GAN architectures for motion prediction, where only GAN losses were used during training. Next, we compared \ourgan-$\ell_2$ to the state-of-the-art trajectory prediction approaches using the full loss with an additional $\ell_2$ loss term, shown in Table \ref{tab:pred-errors_external}.
First, we can see that the introduction of the explicit trajectory loss led to significantly improved performance, as seen when comparing the results of \ourgan from Table \ref{tab:pred-errors} to the \mbox{\ourgan-$\ell_2$} results.
This is consistent with the findings of many other works that using additional task-dependent losses improves the performance of GAN models on supervised learning problems~\cite{gupta2018social, mattyus2018matching}.
In addition, we can see that the $\ell_2$ errors of \mbox{\ourgan-$\ell_2$} are much lower than either \mbox{Social-GAN}~\cite{gupta2018social} or \mbox{Social-LSTM}~\cite{alahi2016social}. 
Both average and final prediction error numbers decreased significantly, showing the benefits of the proposed GAN architecture and the novel differentiable rasterizer.

\subsection{Qualitative results}

In this section we analyze in detail several traffic scenarios commonly seen on the roads, and the outputs of different GAN architectures by sampling $3$ trajectories. 
The scenarios include an actor going straight and turning through an intersection, and an actor approaching an intersection on a straight road.
We show the results in Figure~\ref{fig:case_studies}, comparing \ourgan and \baselinegan trained using only the GAN loss, which only differ in the way the generated trajectories and the rasters are combined at the input of the discriminator (i.e., using differentiable rasterizer or concatenation, respectively).

\begin{table*} [t!]
\centering
\caption{Ablation study of several variants of the proposed \ourgan approach}
\label{tab:ablation-studies}
{\normalsize
{
  \begin{tabular}{lcccccccccc}
     & \multicolumn{6}{c}{\bf mean over 3} & \multicolumn{2}{c}{\bf min over 3} & \multicolumn{2}{c}{\bf min over 20} \\
    \cmidrule(lr){2-7} \cmidrule(lr){8-9} \cmidrule(lr){10-11}
     & \multicolumn{2}{c}{\bf ${\boldsymbol \ell_2}$ [m]} & \multicolumn{2}{c}{\bf ORD [m]} & \multicolumn{2}{c}{\bf ORFP [\%]} & \multicolumn{2}{c}{\bf ${\boldsymbol \ell_2}$ [m]} & \multicolumn{2}{c}{\bf ${\boldsymbol \ell_2}$ [m]} \\
    {\bf Method} & {\bf Avg} & {\bf @4s} & {\bf Avg} & {\bf @4s} & {\bf Avg} & {\bf @4s} & {\bf Avg} & {\bf @4s} & {\bf Avg} & {\bf @4s} \\
    \hline
    \rowcolor{lightgray}
    \ourgan-1channel & 8.68 & 26.52 & {\bf 0.040} & {\bf 0.033 }& {\bf 0.98} & {\bf 1.23} & 8.30 & 26.28 & 8.01 & 25.94 \\
    \ourgan-MNet & 3.82 & 11.18 & 0.723 & 3.068 & 7.21 & 22.43 & 2.62 & 8.15 & 1.82 & 6.08 \\
    \rowcolor{lightgray}
    \ourgan-no-scene & 3.79 & 7.28 & 0.58 & 1.16 & 18.38 & 33.62 & 3.52 & 6.76 & 2.27 & 4.61 \\
    \ourgan & {\bf 2.44} & {\bf 5.86} & { 0.085} & { 0.204} & { 2.11} & { 5.66} & {\bf 1.29} & {\bf 2.95} & {\bf 0.58} & {\bf 1.20} \\
    \hline
\end{tabular}
}
}
\end{table*}

In case \#1 the actor was approaching an intersection in a left-turn-only lane, as indicated by the scene context image.
\baselinegan predicted the actor to keep driving straight, which could be a reasonable prediction considering its past trajectory,
however it is not scene-compliant for this scenario as the predictions entered the lane going in the opposite direction.
On the other hand, the proposed \ourgan correctly predicted the left turn that is compliant with the scene context.
In case \#2 the actor was already inside the intersection, performing an unprotected left turn.
Trajectories output by \baselinegan turned into the wrong lanes (most likely due to potentially too high heading change rate that was reported by the tracker), while trajectories output by \ourgan correctly followed the turning lanes.
In case \#3 the actor was approaching an intersection in a straight-only lane, and
\baselinegan predicted the actor to turn left which is not scene-compliant. 
This happened because the tracked heading slightly tilted to the left, however we can see that \ourgan still correctly predicted the going-straight trajectories.
Lastly, in case \#4 the actor was approaching an intersection in a right-turn-only lane.
While the \baselinegan model incorrectly predicted the actor to keep going straight, the proposed \ourgan correctly predicted the compliant right turn. 
We can see that the proposed model that make use of the differentiable rasterizer outputs trajectories that are much more scene-compliant, leading to better overall prediction accuracy.

\subsection{Ablation study}
In this section we discuss the results of an ablation study of the proposed approach. 
The results are reported in Table~\ref{tab:ablation-studies}, where we compare several variants of the \ourgan model trained using only the GAN loss:
\begin{itemize}
    \item \ourgan-1channel: the rasterized trajectory occupancy grids $[\mathcal{G}^k_{t_c+\delta}, \ldots, \mathcal{G}^k_{t_c+T\delta}]$ are aggregated into a single channel via the \texttt{max} operator instead of being stacked in separate channels;
    \item \ourgan-MNet: the discriminator uses the same CNN network as used in the generator (i.e., using \mbox{MobileNet}~\cite{sandler2018inverted}) instead of DCGAN;
    \item \ourgan-no-scene: the discriminator uses only the trajectory occupancy grids and state inputs, but does not make use of the scene context image.
\end{itemize}

Interestingly, we can see that the \ourgan-1channel model has very low scene-compliance errors, yet high $\ell_2$ errors.
After visualizing its predictions we found the model often predicted static trajectories or trajectories with the points output in random orders.
This was due to the fact that its discriminator is not able to distinguish between two trajectories with the same waypoints but in different orders because of the \texttt{max} aggregation, leading to suboptimal results.
The \mbox{\ourgan-MNet} model had worse metrics than the regular \ourgan model, showing that DCGAN represents a better architecture for the discriminator. 
This is potentially because DCGAN is a more stable architecture as discussed in \cite{radford2015unsupervised}, leading to more useful gradients during training.
Lastly, we observed that the \ourgan-no-scene model also had much worse scene-compliance metrics than the regular \ourgan model. 
This is unsurprising and is similar to the results given in Table \ref{tab:pred-errors}, where we showed that conditioning the discriminator through scene raster input is a critical component for learning improved prediction models.

\section{Conclusion}
\label{sect:conclusion}

Motion prediction is one of the critical components of the self-driving technology, modeling future behavior and uncertainty of the tracked actors in SDV's vicinity. 
In this work we presented a novel GAN architecture to address this task, conditioned on the BEV raster images of the surrounding traffic actors and map elements. 
The main component of the model is the differentiable rasterizer, allowing projection of generated output trajectories into the raster space in a differentiable manner.
This simplifies the task of the discriminator, leading to easier separation of observed and generated trajectories and improved performance of the model.
We evaluated the proposed approach on a large-scale, real-world data set collected by a fleet of self-driving vehicles. 
Extensive qualitative and quantitative analysis showed that the method outperforms the current state-of-the-art in GAN-based motion prediction of the surrounding actors, producing more accurate and realistic trajectories.



\balance 

\bibliographystyle{IEEEtran}
\bibliography{references}

\end{document}